\documentclass[preprint, sort, compress]{elsarticle}

\usepackage{booktabs}       
\usepackage[utf8]{inputenc} 
\usepackage[T1]{fontenc}    
\usepackage{url}            
\usepackage{amsfonts}       
\usepackage{amssymb}
\usepackage{amsmath}
\usepackage{relsize}
\usepackage{bm}

\usepackage[ruled,vlined,linesnumbered]{algorithm2e}

\usepackage{cleveref}
\usepackage{xcolor}
\usepackage{lineno}

\journal{Knowledge-Based Systems}

\makeatletter
\def\ps@pprintTitle{%
 \let\@oddhead\@empty
 \let\@evenhead\@empty
 \let\@oddfoot\@empty
 \let\@evenfoot\@empty
 \def\@oddfoot{\footnotesize Article accepted to Knowledge-Based Systems (2025)\hfill}%
}
\makeatother 

\begin{document}

\begin{frontmatter}



\title{FairUDT: Fairness-aware Uplift Decision Trees \tnoteref{doi}}

\tnotetext[doi]{Published version of the article is available at:
https://doi.org/10.1016/j.knosys.2025.113068.
(https://www.sciencedirect.com/science/article/pii/S0950705125001157)}

\author{Anam Zahid\fnref{myfootnote}}
\ead{anam.zahid@itu.edu.pk}
\author{Abdur Rehman Ali\fnref{myfootnote}}
\ead{abdurrehman.ali@itu.edu.pk}
\author{Shaina Raza\fnref{myfootnote2}\corref{mycorrespondingauthor}}
\ead{shaina.raza@torontomu.ca}
\author{Rai Shahnawaz\fnref{myfootnote}}
\ead{raishahnawaz8@gmail.com}
\author{Faisal Kamiran\fnref{myfootnote}}
\ead{faisal.kamiran@itu.edu.pk}
\author{Asim Karim\fnref{myfootnote3}}
\ead{akarim@lums.edu.pk}
\fntext[myfootnote]{Information Technology University of the Punjab (ITU), Lahore, Pakistan}

\fntext[myfootnote2]{Vector Institute, Toronto Metropolitan University, Toronto, Ontario, Canada}
\fntext[myfootnote3]{Lahore University of Management Sciences (LUMS), Lahore, Pakistan}

\cortext[mycorrespondingauthor]{This is to indicate the corresponding author.}

\begin{abstract}
Training data used for developing machine learning classifiers can exhibit biases against specific protected attributes. Such biases typically originate from historical discrimination or certain underlying patterns that disproportionately under-represent minority groups, such as those identified by their gender, religion, or race. In this paper, we propose a novel approach, FairUDT, a fairness-aware Uplift-based Decision Tree for discrimination identification. FairUDT demonstrates how the integration of uplift modeling with decision trees can be adapted to include fair splitting criteria. Additionally, we introduce a modified leaf relabeling approach for removing discrimination. We divide our dataset into favored and deprived groups based on a binary sensitive attribute, with the favored dataset serving as the treatment group and the deprived dataset as the control group. By applying FairUDT and our leaf relabeling approach to preprocess three benchmark datasets, we achieve an acceptable accuracy-discrimination tradeoff. We also show that FairUDT is inherently interpretable and can be utilized in discrimination detection tasks. 



\end{abstract}


\end{frontmatter}



\section{Introduction}
\label{intro}
Discrimination represents the unfair treatment of individuals based on their association with specific groups, influenced by characteristics like nationality, sex, age, and race. Globally, laws exist to combat such discrimination in various sectors, including employment, education, and housing \citep{usgovernment2011, barocas2016big}. However, machine learning (ML) classifiers, often trained on historical data, may perpetuate existing biases if the data itself is biased \citep{calders2013unbiased}.

To mitigate this, algorithmic bias countermeasures can be classified into pre-processing, in-processing, and post-processing categories. Pre-processing methods modify the datasets to eliminate bias \citep{ luong2011k, calmon2017optimized, feldman2015certifying, madras2018learning}. In contrast, in-processing techniques adjust the model's learning mechanisms \citep{raff2018fair, kamiran2010discrimination, aghaei2019learning, valdivia2020fair}, and post-processing methods focus on altering the predictions \citep{Lohia2019, hardt2016equality, awasthi2020equalized}.

Traditional classifiers like naïve bayes and logistic regression predict probabilities based on attributes, while uplift modeling (a specialized area of predictive analytics) aims to measure the impact of interventions on {\it treatment} vs {\it control} groups \cite{gutierrez2017causal}. Specifically, we could define the treatment group as those subjected to an action or intervention (e.g., favoritism), while the control group remains deprived of this phenomenon. This method is particularly effective for identifying discrimination by comparing the probability differences between favored and deprived groups, which helps in pinpointing instances of undue favoritism or deprivation.

Building on these concepts, we introduce FairUDT (Fairness-aware Uplift Decision Trees) , a novel decision tree model designed to identify and address biases in decision-making processes.  To our knowledge, FairUDT is the first method that employs uplift modeling for this purpose. Our approach modifies decision trees by incorporating a unique leaf relabeling criterion that targets only discriminatory instances instead of entire groups, enhancing fairness while maintaining accuracy.

\subsection{Research Objectives:}
The objectives of this study are:
\begin{itemize}

\item To explore the use of uplift modeling within decision trees as a method for identifying and addressing biases against marginalized groups.
\item To develop and evaluate FairUDT, a novel approach that combines uplift modeling and decision trees for reducing bias, with broad applicability across various machine learning contexts.
\item To propose new fairness-aware decision tree construction techniques, including different criteria for splitting nodes and a refined leaf relabeling approach.
\item To study the interpretability of FairUDT, ensuring it remains a transparent and user-friendly tool for bias detection and mitigation.
\item To assess the effectiveness and transparency of FairUDT through comparative analysis with existing methods, using a range of fairness and accuracy metrics.
\end{itemize}

\subsection{Our Contributions:}
The main contributions of this work are summarized as follows:
\begin{itemize}
\item  We pioneer the application of uplift modeling to discrimination identification, providing a novel analytical framework for fair machine learning.
\item We propose FairUDT, an uplift modeling based technique for fairness-aware decision tree construction and, a selective leaf relabeling approach. This methodology enables precise discrimination removal by targeting only those subgroups exceeding specific discrimination thresholds, rather than entire leaves.
\item  Our model is rigorously tested on three benchmark datasets—Adult \citep{AdultDataset}, COMPAS \citep{COMPASSDataset}, and German Credit \citep{dua2017uci}—showing robust performance across all, with superior results on COMPAS and German Credit datasets. This demonstrates our method's effectiveness and generalizability, compared with existing dataset-specific techniques.
\item The proposed model enhances interpretability, allowing stakeholders to easily understand and address potential subgroup discriminations within decision-making processes. We make the method and code \footnote{\url{https://github.com/ara-25/FairUDT}} available for the reproducibility of research.
\end{itemize}


\section{Related work}
This section discusses the most recent research in algorithmic discrimination, as well as the application of decision trees in uplift modeling. It also highlights how our proposed work connects these two domains. 

\textbf{Algorithmic discrimination :} 
\label{lit-review-disc}
Traditional pre-processing based techniques aim to produce “balanced” datasets that can be fed into any machine learning model. These techniques can be categorized as relabelling, resampling, and data transformation techniques. In relabelling, only the target values of a few training samples are changed \citep{kamiran2009classifying, luong2011k}. Resampling involves duplicating or dropping specific samples, or assigning weights, as in the reweighing method proposed by \citet{kamiran2012data}. Data transformation techniques manipulate the training features and their labels to generate new training sets. \citet{calmon2017optimized} propose an `optimized pre-processing algorithm' to minimize discrimination while controlling the distortion and utility in the resulting transformed data. However, the optimization programs can be computationally very expensive to solve. Disparate impact remover \citep{feldman2015certifying} is another data transformation technique, focused on satisfying the disparate impact fairness criteria, also known as the 80\% rule \citep{usgovernment2011}. It applies a repair procedure that removes the relationships between protected attribute and other input features. Recently, pre-processing approaches based on adversarial learning have also been developed \citep{madras2018learning}. However, these techniques are complex, require more computation power, and are difficult to interpret in terms of the resulting representations. In contrast, our proposed approach is focused on identification and relabeling of discriminatory subgroups in the data by building interpretable decision trees and is thus more intuitive.

In the domain of discrimination aware data mining, many in-processing approaches are also proposed for bias mitigation. A recent study analyzed fairness in in-processing algorithms \citep{wan2023processing} and divided these mitigation approaches into two categories: explicit and implicit mitigation methods. Explicit mitigation refers to modification of the objective function by either adding a regularization term \cite{jiang2020wasserstein, aghaei2019learning, agarwal2019fair} or by constraint optimization \citep{Saxena2024, garcia2021maxmin, lahoti2020fairness, garg2019counterfactual}. Implicit methods aim to improve latent representations by learning adversarial \citep{sweeney2020reducing, zhang2018mitigating}, disentangled \citep{kim2021counterfactual, park2021learning} and contrastive \citep{cheng2020fairfil, zhou2021contrastive} representations. Discrimination aware post-processing techniques, on the other hand, modify the output of a standard learning algorithm by using some non-discriminatory constraints \citep{kamiran2012icdm, hardt2016equality, Lohia2019, awasthi2020equalized}.

\textbf{Discrimination-aware decision trees:} Decision trees are a widely used supervised machine learning algorithm for classification and regression tasks. They have a hierarchical structure with branches, nodes, and leaves. The algorithm splits the dataset at each internal node based on decision rules from data attributes, leading to leaf nodes with the final classification or regression outcomes.

In traditional decision trees, each split aims to maximize information gain, resulting in branches with more homogeneous subsets of data, thereby improving prediction accuracy. However, in discrimination-aware decision trees, the objective is to balance accuracy with fairness. This involves a multi-objective optimization at each split, ensuring that outcomes are equitable across sensitive groups. The concept was first introduced by \citet{kamiran2010discrimination}, who modified the splitting criteria and applied a leaf relabeling approach to achieve non-discriminatory outcomes.
Recent approaches in constructing discrimination aware decision trees are more focused on modifying existing splitting criterion to account for fairness besides maximizing predictive performances. For example, \citet{raff2018fair} uses the difference method, originally proposed by \cite{kamiran2010discrimination}, to calculate information gain in CART decision Trees. Similarly, work by \citet{aghaei2019learning} added fairness constraints to the loss function of CART trees using Mixed-Integer Programming (MIP) model. The work of \cite{valdivia2020fair} involves optimizing multiple objective functions of decision trees using genetic algorithms. Work by \cite{zhang2020feat} propose modified hoeffding trees for maintaining fairness in streaming data. A more recent study introduced FFTree \citep{castelnovo2022fftree}, an in-processing algorithm that uses decision trees to select multiple fairness criteria and sensitive attributes at each split. The goal of FFTree is to select best split where information gain is optimal with respect to both fairness criteria and sensitive attributes. However, because of their flexibility in picking multiple fairness criteria and sensitive features, FFTree may suffer greater losses in accuracy than typical discrimination-aware decision trees.  

\textbf{Uplift modeling :} 
Most work in the domain of machine learning and Causal Inference (CI) has been focused on structural causal modeling i.e. learning causal graphs and counterfactuals. Another goal of CI is to measure the effect of potential cause (e.g. event, policy, treatment) on some outcome. Uplift modeling is associated with this potential outcomes framework of CI modeling.
The major focus of uplift modeling is to measure the effect of an action or a treatment (e.g. marketing campaign) on customer outcome \citep{gutierrez2017causal}. The term “uplift” specifically refers to estimating the differences between the buying behavior of customers given the promotion (treatment) and those without it (control) \citep{ jaskowski2012uplift}. Current uplift modeling techniques can be distributed in three categories: tree based uplift models and its ensembles, SVM based models and generative deep learning methods. While generative deep learning models are more recently introduced, we will continue our discussion on tree based uplift models since they are the main focus of our approach. Most of the existing tree based uplift modeling strategies used modified splitting criteria for measuring treatment effect \citep{gaoutboost2023}. For example, uplift incremental value modeling \citep{hansotia2002incremental} and uplift t-statistics tree \citep{Su2009subgroup} tends to maximize the splitting criteria by measuring conditional average treatment effect between left and right child node of a tree. However, work by \citet{rzepakowski2010decision} tries to maximize the difference of average outcome between control and treatment groups within a single decision tree node.

To the best of our knowledge, FairUDT is the first paper to study uplift modeling in terms of discrimination identification. A similar approach in the literature is proposed \citep{he2020inherent}, where causal trees are used for the discovery of discriminatory subgroups and the tree splitting criteria is based on measuring Heterogeneous Treatment Effect (HTE). A limitation of using causal trees is that it requires consistency estimates i.e. balanced number of favored and deprived individuals in the leaf nodes \citep{Athey7353}. This is practically impossible in bias quantification since deprived individuals are present in minorities mostly. Our approach, on contrary, deals with measuring differences in probability distribution between favored and deprived groups and does not require any such assumption.

 \textbf{Discrimination identification using uplift modeling :} Discrimination can be defined as the {\it class probabilities difference} among favored and deprived distributions for the groups exhibiting the same characteristics or features. This notion of discrimination identification is aligned with the idea of uplift modeling where \textbf{incremental impact} for treatment group was differentiated from control group \citep{rzepakowski2010decision}. In a marketing campaign, given a treatment and control set of users, uplift is measured as the difference between response rates for the two sets. Making use of the same analogy for a conventional discriminatory setting, the \textbf{disparate impact} between the favored and deprived groups can be measured by considering the difference between the acceptance and rejection rates among two groups. For example, given similar qualifications among members of both groups applying for the same job, how much is one group preferred over other.

 \textbf{Interpretable Decision Trees :} In the field of machine learning, interpretability is defined as the ability to explain models and their outputs in a manner that is comprehensible to humans. However, quantitative notions of interpretability are largely absent from the machine learning literature, complicating the comparison of models without human intervention \cite{aghaei2023optimalfair}. One such measure is sparsity, which refers to the simplicity of models within the same class. In the context of decision trees, \citet{rudin2022interpretable} defines sparsity as the number of leaves in the tree, where trees with fewer leaves are considered sparser and therefore, more desirable. Another important measure is simulatability \cite{Scarpato2024}, i.e. the ease of the humans to interpret part or all of the model decisions. Shallow trees are known to be most simulatable as the if-else decision rules generated by these trees are easily understandable by humans. Some studies \cite{molnar2022, carvalho2019machine} suggested that the depth of the decision tree can also be taken as a complexity indicator, where more depth means more complex model. 

\section{FairUDT - Fairness-aware Uplift Decision Trees}

In this study, we propose FairUDT, a novel decision tree model designed to identify and address biases in decision-making processes. FairUDT employs an uplift-induced splitting criterion that quantifies the {\it class probabilities difference} between favored and deprived groups at each node of the tree. This criterion helps to detect discriminatory patterns, referred to as uplifts, in the model's predictions.

The tree is systematically developed until all leaves display maximal probabilistic differences, signifying potential uplifts (or biases). To mitigate these biases, we introduce a fairness-oriented intervention through a leaf relabeling technique. This approach adjusts the labels of data points within biased subgroups based on their discrimination scores, aiming to achieve equitable treatment across groups. The end-to-end pipeline of our proposed fairness-aware approach, FairUDT and leaf relabeling, is shown in \Cref{prepipe}.

\subsection{Model Construction}
The construction of FairUDT is different from traditional decision tree models by utilizing dual datasets that represent both favored and deprived groups. These groups are defined based on a sensitive attribute, allowing the tree to simultaneously evaluate and adjust for biases during its construction. This section delineates various splitting criteria implemented to enhance fairness in the decision-making process. Further details on the methodologies for identifying biased sectors and the specific leaf relabeling techniques used to treat these biases are discussed in Section \ref{leaf_relab}. This approach ensures that FairUDT not only identifies but also rectifies biases, promoting fairness in predictive modeling.  


\subsection{Notations and Definitions}
\label{notation}
Given a dataset \((X, S, Y)\), with \(N\) records, where \(X\) represents non-sensitive attributes and \(S = \{S_F, S_D\}\) is a binary sensitive attribute like gender or race. The dataset is split into favored \((X, S_F, Y)\) and deprived \((X, S_D, Y)\) groups, containing \(N^F\) and \(N^D\) records, respectively. The outcome \(Y\) is binary \(\{Y^+, Y^-\}\).

For categorical attributes, we perform a split test \(A\) with outcomes \(a \in A\), where numerical attributes are discretized. Conditional probabilities are calculated as \(P^F(Y|a)\) and \(P^D(Y|a)\) for each outcome \(A=a\) in the favored and deprived groups. A summary of the notations used in this paper is given in Table \ref{tab:notations}.

\begin{table*}[t]
\centering
\begin{tabular}{cl}
\hline
\textbf{Notation} & \textbf{Description} \\
\hline
\(X\) & Non-sensitive attributes \\
\(S\) & Binary sensitive attribute \(\{S_F, S_D\}\) such as gender or race \\
\(Y\) & Binary outcome variable \(\{Y^+, Y^-\}\) \\
\(N\) & Total number of records in the dataset \\
\(N^F\) & Number of records in the favored group \\
\(N^D\) & Number of records in the deprived group \\
\(P^F\), \(P^D\) & Probabilities for the favored and deprived groups \\
\(P^F(Y)\), \(P^D(Y)\) & Probabilities of the outcome \(Y\) within the favored/deprived groups \\
\(A\) & Categorical attribute with outcomes \(a\) \\
\hline
\end{tabular}
\caption{Summary of notations used in the analysis.}
\label{tab:notations}
\end{table*}

We define Demographic Parity (DP) as equal probability of a positive outcome across groups:
    \[
    DP = | P(\hat{Y} = Y^+ | S = S_F) - P(\hat{Y} = Y^+ | S = S_D) |
    \]
    A value of 0 indicates perfect parity.
   
We define the Average Odds Difference (AOD) as fairness metrics as the mean of the differences in True Positive Rates (TPR) and False Positive Rates (FPR) between groups:
    \[
    AOD = \frac{[TPR_{S = S_F} - TPR_{S = S_D}] + [FPR_{S = S_F} - FPR_{S = S_D}]}{2}
    \]
    An AOD of 0 signifies no disparity in odds.

We also employ Balanced Accuracy (BA) to evaluate classifier performance, calculated as the average of TPR and True Negative Rate (TNR):
\[
BA = \frac{TPR + TNR}{2}
\]

The details are given in \ref{notationsAppendix}

\subsection{Uplift modeling based splitting criteria}
Classical splitting criteria for decision tree construction focus on modeling class probabilities to ensure accurate predictions. However, fairness in decision-making represents a Pareto optimization problem, requiring decisions to be both accurate and fair.

In standard decision tree construction, an attribute test yielding the maximum information gain for class {\it (IGC)} is selected for the split. In the seminal work of \cite{kamiran2010discrimination}, multiple splitting criteria were designed to minimize bias in decision-making. Alongside {\it IGC}, the information gain with respect to the sensitive attribute {\it (IGS)} is also calculated for the split tests. The intuition behind this approach was to decide splits based on the maximum difference between {\it IGC} and {\it IGS} (i.e., {\it IGC-IGS}), aiming to produce leaves that are homogeneous with respect to the class label and heterogeneous with respect to the sensitive attribute. However, this approach and the {\it IGC/IGS} ratio did not demonstrate significant improvements in reducing discrimination. Another proposed criterion, {\it (IGC+IGS)}, aimed at achieving homogeneity with respect to both class and sensitive attributes, yet it often lacked variation in the sensitive attribute to effectively adjust for fairness. Nevertheless, as results for such splits, when combined with a leaf relabeling technique, were better, it was adopted.

In contrast, the splitting criteria of FairUDT model the {\it difference in class probabilities} between favored and deprived groups instead of modeling the actual class probabilities, with the aim of identifying subgroups at leaves that maximize uplift, or in our case, discrimination. As a result, biased sectors are identified among the tree leaves and can be addressed for fairness later. This approach is also conventional to decision tree construction as it models the amount of {\it information} a test provides about this difference. Given the goal to optimize the probabilistic differences between favored and deprived groups, our splitting criterion is logically based on distribution divergence.

Following the work of \cite{rzepakowski2010decision}, we use two different distribution divergence measures: Kullback-Leibler divergence (KL) and squared Euclidean distance. KL-divergence \citep{csiszar2004information} is a widely recognized information-theoretic measure. Although Euclidean distance is less commonly applied for comparing probability distributions, it is referenced in the literature \citep{soltys2015ensemble}. Given two separate probability distributions \(Q=(q_1, q_2, \ldots, q_n)\) and \(P=(p_1, p_2, \ldots, p_n)\), the divergences are defined as follows:

\begin{equation}
\label{KL} 
KL(P:Q)= \sum_{i=1}^n p_i \log{\frac{p_i}{q_i}} \tag{3.1}
\end{equation}

\begin{equation}
\label{Euclid}
E(P:Q)= \sum_{i=1}^n {(p_i-q_i)}^{2} \tag{3.2}
\end{equation}

For a given split test \(A\), the gain or increase in divergence for any divergence measure \(D\) would be the difference between divergence values after and before the split:

\begin{equation}
\label{divergenceGain}
D_{gain}(A) = D(P^F(Y):P^D(Y)|A) - D(P^F(Y):P^D(Y)) \tag{3.3}
\end{equation}

The first part of the equation denotes the conditional divergence among favored and deprived distributions after performing a split test on an attribute \(A\), whereas the second part measures divergence prior to the split. $D_{gain}$ quantifies the overall change in divergence resulting from the split on test \(A\). A positive gain signifies that the split has increased the divergence between favored and deprived groups. Conversely, a negative gain indicates that the split on \(A\) has decreased the divergence, suggesting that \(A\) is a less significant candidate for tree construction.

Conditional divergences, used multiple times in the literature \citep{han2002mathematics,yu2020measuring}, are adapted for our setting as defined by \cite{rzepakowski2010decision}, taking into account two probability distributions separately (i.e., favored and deprived in our case).

\begin{equation}
\label{divergenceTest}
D(P^F(Y):P^D(Y)|A) = \sum_{a} {\frac{N(a)}{N}} D(P^F(Y|a):P^D(Y|a)) \tag{3.4}
\end{equation}

Here, \(\sum_{a}\) sums over all outcomes of test \(A\), weighting each specific test value by the fraction of instances for that respective test outcome, analogous to definitions of {\it gini} and {\it entropy} measures.

\subsubsection{Kullback-Leibler divergence}
Kullback-Leibler divergence is a directed divergence which measures an asymmetric distance between two probability distributions. Kullback preferred the term "discrimination information" \citep{kullback1987letter}. The asymmetry of KL-divergence is not problematic in our context, as the deprived group serves as an inherent reference point from which the favored group is expected to diverge due to its preferential treatment. In fact, our experiments confirm that KL-divergence is a more appropriate measure for our scenario, as it is sensitive to distributional differences and effectively captures the directional information of divergence. Mathematically, gain in discrimination for a split test is computed by substituting conditional divergence \Cref{divergenceTest} in \Cref{divergenceGain}, replacing D with KL and finally substituting KL value from \Cref{KL}.

\begin{equation}
\label{KLdivergenceGain}
KL_{gain}(A) = \mathlarger{\mathlarger{\sum_{a}}} \left( \frac{N(a)}{N} \sum_{y} f_y(a) \log{\frac{f_y(a)}{d_y(a)}} \right) - \sum_{y} f_y \log{\frac{f_y}{d_y}} \tag{3.5}
\end{equation}

Here, \(y\) denotes a particular class label, \(\sum_{y}\) is the sum over all class labels, and \(f_y\) and \(d_y\) depict the probability of \(y\) in the favored and deprived distributions, respectively. Similarly, \(f_y(a)\) and \(d_y(a)\) denote the corresponding probabilities for a certain test outcome \(a\). We used Laplace correction throughout while estimating the probabilities \(P^F\) and \(P^D\) because the KL value will tend to infinity if, for any test outcome, \(d_y(a)\) is zero and the corresponding \(f_y(a)\) is non-zero. We establish some facts through propositions which $KL_{gain}$ should satisfy (see \ref{KLAxiomAppendix}{}).

\paragraph{$\boldsymbol{KL_{gain}}$ Ratio}
Standard decision tree learners address the bias of information gain on high-branching attributes using the split information for the test. In our case, at first, the split information for favored and deprived groups can differ. Additionally, we need to account for uneven splits between favored and deprived distributions to ensure randomness and avoid problems with probability estimation. In an extreme case, a test can split the favored and deprived groups into different subtrees, making it difficult to detect discrimination. As a result, the subsequent tree construction for either the favored or deprived group will rely on conventional entropy gain. Hence, the suggested normalization value for the $KL_{gain}$ test is as follows:
(Recall $N=N^F+N^D$ denotes the total number of records for favored and deprived datasets.)

\begin{equation}
\label{KLnormalizationValue}
\begin{split}
     I_{KL}(A)= H\left(\frac{N^F}{N},\frac{N^D}{N}\right)KL(P^{F}(A):P^{D}(A)) \\
+ \frac{N^F}{N}H(P^{F}(A)) + \frac{N^D}{N}H(P^{D}(A)) 
\end{split}
\tag{3.6}
\end{equation}
The first term of the equation punishes the tests that split the favored and deprived groups into different proportions. The imbalance between proportions is calculated using the divergence between two distributions, which can be arbitrarily close to infinity. However, it makes no sense to penalize on this account when there is not sufficient data available for favored and deprived groups. Therefore, the divergence part is multiplied by the entropy function \(H\left(\frac{N^F}{N},\frac{N^D}{N}\right)\), which approaches zero for massive disproportion between the number of data records for the involved groups. The second and third terms of the equation add split information for favored and deprived groups as a fraction of their records involved, hence penalizing tests with a large number of outcomes. One more concern is that very small values of the normalizing factor can inflate the value of tests with even low $KL_{gain}$. To resolve this, a test is selected only if it has a greater or equal value to the average gain of all attributes. The final splitting criterion would be the fraction of \Cref{KLdivergenceGain} and \Cref{KLnormalizationValue}.

\begin{equation}
KL_{gain} Ratio = \frac{KL_{gain}}{I_{KL}(A)} \tag{3.7}
\end{equation}

\subsubsection{Squared Euclidean Distance}
Euclidean distance was considered because of its certain advantages over KL-divergence as a distance measure. Unlike KL, it is symmetric, which has benefits in tree learning when the favored group is missing. Moreover, it is more stable as it does not require Laplace correction for the probabilities to avoid infinitely large uncertain values that may lead to the wrong selection of a test attribute. Euclidean divergence gain for a test over distributions \(P^F\) and \(P^D\) is defined as follows:

\begin{equation}
\label{EuclideanGain}
E_{gain}(A)= \mathlarger{\mathlarger{\sum_{a}}} \left( \frac{N(a)}{N} \sum_{y} (f_y(a)-d_y(a))^2 \right) - \sum_{y} (f_y-d_y)^2 \tag{3.8}
\end{equation}

The notation description is analogous to that defined for \(KL_{gain}\) in \Cref{KLdivergenceGain}. Few claims for $E_{gain}$(A) vital to our context are also discussed (see \ref{EucledianAxiomAppendix}).

\paragraph{$\boldsymbol{E_{gain}}$ Ratio}
 
The explanation for the suggested normalization value for the \(E_{gain}\) test is analogous to that used for \(KL_{gain}\) in \Cref{KLnormalizationValue}. The only exception is the use of the Gini calculation instead of entropy for all the terms involved to ensure symmetry. 

The first term of the equation penalizes tests that create disproportionate splits between favored and deprived groups, measured by squared euclidean divergence. However, this penalty is adjusted by the function \(GINI\left(\frac{N^F}{N},\frac{N^D}{N}\right)\), which approaches zero when there is a significant disparity in data records between the groups.
The second and third terms incorporate split information for both groups based on the fraction of records involved, penalizing tests with many outcomes. This is done using the sum of Gini indices for the test’s outcomes, weighted by the total number of records in each group.

\begin{equation*}
\label{EnormalizationValue}
\begin{aligned}
I_E(A)= GINI\left(\frac{N^F}{N},\frac{N^D}{N}\right)E(P^F(A):P^D(A)) \\
+ \frac{N^F}{N}GINI(P^F(A)) + \frac{N^D}{N}GINI(P^D(A))
\end{aligned}
\tag{3.9}
\end{equation*}

The final splitting criterion would be formulated as follows:

\begin{equation}
E_{gain} Ratio = \frac{E_{gain}}{I_E(A)} \tag{3.10}
\end{equation}

\subsection{Discriminatory Regions and Tree Implications}
Subsequent to FairUDT construction, discrimination can arise at a certain leaf \(l \in L\) if the probability of the positive class for the favored subgroup is higher than the probability for the deprived subgroup, i.e., \(P^F(y^+|l) > P^D(y^+|l)\) (similarly, the probability of the negative class in the deprived subgroup is greater than the probability in the favored subgroup, i.e., \(P^D(y^-|l) > P^F(y^-|l)\)). Since FairUDT is built using probability divergence measures, the resulting leaves can contain discrimination information for subgroups, making it readily available for mitigation. Formally, the overall discrimination at a certain leaf \(disc_l\) can be calculated as the sum of discrimination values for the positive and negative classes:

\begin{equation}
\label{discrimination}
disc_l = (P^F(y^+|l) - P^D(y^+|l)) + (P^D(y^-|l) - P^F(y^-|l)) \tag{3.11}
\end{equation}
\\
\textbf{Definition 1:} A leaf \(l\) is said to be a discriminatory region if \(disc_l > 0\).

After identifying the discriminatory regions, leaf relabeling is applied either by demoting the favored individuals or by promoting the deprived individuals, and the resultant decision tree can be adapted as a data pre-processor. Furthermore, a tunable parameter \(\sigma_t\) is adjusted to select leaves for relabeling.
\\
\textbf{Definition 2:} A leaf \(l\) can only be relabeled if \(disc_l > \sigma_t\).

The selection of the tunable parameter \(\sigma_t\) depends on business needs, as there is no single standard for allowable fairness in the industry. Moreover, tuning \(\sigma_t\) to achieve a completely fair dataset may result in poor performance, i.e., a tradeoff between fairness and performance \citep{castelnovo2022fftree}. Hence, a domain expert is required to determine the optimal value of \(\sigma_t\) that minimizes loss of accuracy. In our settings, the value of \(\sigma_t\) ranges from 0 to 2. A value of \(\sigma_t = 0\) indicates that all instances in the leaf must be relabeled.

\begin{figure*}[htbp]
 \centering
\includegraphics[width=1\linewidth]{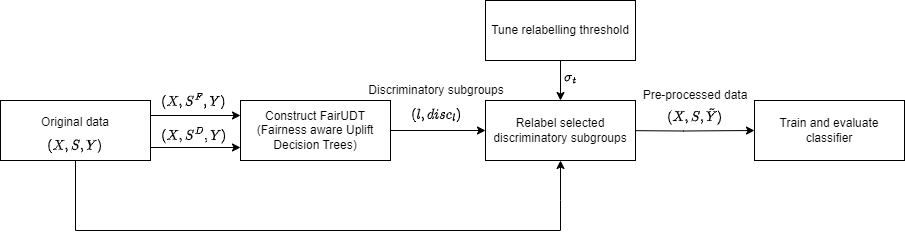}
  \caption{End-to-end pipeline of FairUDT and leaf relabelling}
  \label{prepipe}
\end{figure*}

\section{Leaf Relabeling}
\label{leaf_relab}

Contrary to the definition in \Cref{discrimination}, the work by \cite{kamiran2010discrimination} defined discrimination separately for positively and negatively labeled leaves, which is simply the difference in fractions of records between favored and deprived subgroups at a certain leaf. Hence, for leaves with a positive majority class, discrimination was quantified as \(P^F(Y|l) - P^D(Y|l)\), whereas leaves with a negative majority class had a discrimination value equal to \(P^D(Y|l) - P^F(Y|l)\). However, this formulation of discrimination is applicable only in reference to the majority class. In general, it is terribly faulty on multiple accounts. Consider a leaf having similar proportions of favored and deprived instances, but with favored instances having positive and deprived instances having negative actual class labels. This should be considered a highly discriminatory region since, in a leaf exhibiting similar features, the class probabilities differences (i.e., both acceptance and rejection rates differences) are maximal between favored and deprived subgroups. Entirely opposed to that, the discrimination value would be zero as per their definition. Thus, the settled concept for quantifying discrimination at leaves is strictly contextual.

In preference to the previous approach, our proposed leaf relabeling approach is based on demoting the unnecessarily favored individuals or promoting the unjustifiably deprived individuals. Intuitively, if the majority of a subgroup is accepted, then lower acceptance and higher rejection rates for the deprived demographic within that subgroup are unjustified. Similarly, if the majority of a subgroup is rejected, then lower rejection and higher acceptance rates for the favored group within this division also indicate disparity. Henceforth, we have devised a strategy to remove disparate impact by relabeling randomly selected individual records at a certain discriminatory leaf {\it l} such that the discrimination defined by \Cref{discrimination} becomes 0, i.e., the equality conditions \(P^F(y^+|l) = P^D(y^+|l)\) and \(P^F(y^-|l) = P^D(y^-|l)\) are satisfied. This is done by promoting or demoting individual records at that leaf to the majority class.\\
\textbf{--Promotions:} If the dominant class at a discriminatory leaf is positive, then random deprived individuals at that leaf with a negative class are promoted in the dataset. \\
\textbf{--Demotions:} If the majority class at a discriminatory leaf is negative, then random favored individuals at that leaf are demoted in the dataset. For equal proportions, promotions to the positive class are preferred. The complete procedure to generate the relabeled dataset \((X,S,\Tilde{Y})\) is shown in Algorithm \ref{alog:preprocess}.

In contrast to most previous works on pre-processing \citep{calmon2017optimized, hajian2013methodology, ruggieri2014using}, our relabeling approach incentivizes {\it utility} preservation by promoting or demoting individuals to the majority class.

The end-to-end pipeline of our proposed fairness-aware approach, FairUDT and leaf relabeling, is shown in \Cref{prepipe}. The favored and deprived groups from the raw data \((X, S, Y)\) are first utilized in building FairUDT. After the construction of the tree, the biased subgroups are identified among the leaves \(L\) of the tree, where the discrimination \(disc_l\) in each leaf \(l\) is quantified by \Cref{discrimination}. Once a tree is built, all leaves having discrimination greater than \(\sigma_t\) are relabeled. Finally, the pre-processed data \((X, S, \Tilde{Y})\) is used to train a classification algorithm. It is important to note that the test data must also be pre-processed before making predictions with the classifier.
 
 \begin{algorithm}[!htpb]
\label{alog:preprocess}
\linespread{0.9}\selectfont

\DontPrintSemicolon
\SetAlgoLined
\SetKwFunction{FPromote}{Promote}
\SetKwFunction{FDemote}{Demote}
\SetKwInOut{Input}{Input}\SetKwInOut{Output}{Output}
\SetKwFunction{non_empty}{non_empty}
\Input{Original data $(X,S,Y)$, the discriminatory subgroups on leaves $L$, data instances $(X_l,S_l,Y_l)$ for every leaf $l \in L$, and discrimination threshold $\sigma_t \in [0,2]$}

\Output{Pre-processed data $(X,S,\Tilde{Y})$}
\BlankLine
Initialize $\Tilde{Y} \leftarrow Y$
\BlankLine
\ForEach{\{$l \in L\}$}{
    Compute $disc(l)$ as in Eq. (\ref{discrimination})\;
    Compute $class(l) := mode(Y_l)$ \tcc*[l]{Find majority class in the subgroup $l$}
}
\ForEach{\{$l \in L | disc(l) \geq \sigma_t\}$}{
    \lIf{$class(l) = y^+$}{
        \FPromote{$\Tilde{Y}$, $l$};
    }
    \lElse{
        \FDemote($\Tilde{Y}$, $l$);
    }
}

\BlankLine

\Return{$\Tilde{Y}$}

\caption{Leaf relabeling}
\BlankLine
\SetKwProg{Fn}{Function}{:}{}
\Fn{\FPromote{$\tilde{Y}$, $l$}}{
    \tcc{select deprived negative instances from leaf \textit{l} which will be relabeled}
    Initialize $R \leftarrow \tilde{Y}[h]$ where $h = \{e|(e \in Y_l) \wedge (S_l=s_D) \wedge (Y_l=y^-)\}$\; 
    \tcc{Find the number of deprived negative instances to relabel such that the equality conditions $P(y^+|s_F)$ = $P(y^+|s_D)$ and $P(y^-|s_F)$ = $P(y^-|s_D)$ are satisfied in this subgroup}
    $p = \lfloor{P(y^+|s_F)P(s_D) - P(y^+|s_D)}\rfloor$\;
    Initialize $T \leftarrow p$ random elements from $R$\;
    \ForEach{$t \in T$}{
        $t := y^+$ \tcc*[l]{Relabel instance to the positive class}
    }
}
\BlankLine
\Fn{\FDemote{$\tilde{Y}$, $l$}}{
    \tcc{select favored positive instances from leaf \textit{l} which will be relabeled}
    Initialize $R \leftarrow \tilde{Y}[h]$ where $h = \{e|(e \in Y_l) \wedge (S_l=s_F) \wedge (Y_l=y^+)\}$\; 
    \tcc{Find the number of favored positive instances to relabel such that the equality conditions $P(y^+|s_F)$ = $P(y^+|s_D)$ and $P(y^-|s_F)$ = $P(y^-|s_D)$ are satisfied in this subgroup}
    $p = \lfloor{P(y^-|s_D)P(s_F) - P(y^-|s_F)}\rfloor$\;
    Initialize $T \leftarrow p$ random elements from $R$\;
    \ForEach{$t \in T$}{
        $t := y^-$ \tcc*[l]{Relabel instance to the negative class}
    }
}
\end{algorithm} 

\begin{table}[t]
\caption{Statistics of the datasets}
\label{tb:data}
\centering
\smaller
\fontsize{6pt}{8pt}\selectfont

\begin{tabular}{c|c|c|c|c|c}
\toprule
\textbf{Dataset}  & $N$ & $N^F$ & $N^D$ & S = {[} ${S}_F , {S}_D {]} $ & Attributes \\
\midrule
Adult & 45,222 & 30,527 & 14,695 & gender = {[}male, female{]} & 14 \\
\hline
COMPAS & 6,167 & 2,100 & 4,067 & race ={[}caucasian, non-caucasian{]}&  9 \\
\hline
German Credit  & 1,000 & 810 & 190 & age ={[}\textgreater{}25, \textless{}=25{]} & 20  \\
\bottomrule
\end{tabular}
\end{table}

\section{Experimental Setup}
\subsection{Datasets}
We evaluated our approach on three real-world datasets, utilizing FairUDT as the data pre-processor. The summary statistics of these datasets are given in Table \ref{tb:data}.
The first one, UCI Adult dataset \citep{AdultDataset} was used to determine  whether the annual income of a person is greater than 50k dollar or not. The second dataset, ProPublica COMPAS dataset \citep{COMPASSDataset} evaluates whether or not an individual will be arrested again within two years of its release. The individual's {\it race} is used as the sensitive attribute and has been binarized to {\it caucasian} and {\it non-caucasian} categories. The UCI German Credit dataset \citep{dua2017uci} predicts if an individual has good or bad credit risk. 

\subsection{Hyperparameters and Settings}
For the evaluation of the proposed uplift-based discrimination-aware decision tree FairUDT, we used all the relevant dataset attributes during the pre-processing and classification phases. Using a larger group of features with more categories per feature incentivizes the discovery of fine-grained subgroups. Furthermore, the numerical features were quantized to categorical ranges.

We did not implement any early stopping methodology; thus, FairUDT was extended to its maximum depth to capture highly biased instances at the tree leaves. We conducted experiments using both the ${KL}_{gain}$ and ${E}_{gain}$ ratio splitting criteria for the construction of FairUDT.

Following the pre-processing of the original dataset, we employed it to fit Logistic Regression (LR), Decision Tree (DT), and Support Vector Machine (SVM) classifiers without any hyperparameter tuning, utilizing a 75-25 train-test split. All classification experiments were performed using 10-fold cross-validation and the results are reported on both the original and relabeled test sets. Additionally, to compare our results with key related prior works, we utilized the pre-configured notebooks from AIF360, an open-source toolkit for fairness-aware machine learning \cite{bellamy2018ai}. We used the default hyperparameter values for the previously proposed methods to ensure an unbiased comparison of results under consistent conditions.

\section{Results}

In this section, we first present the results of the comparative analysis of FairUDT with key related prior works on data pre-processing: \textbf{optimized pre-processing} \citep{calmon2017optimized}, \textbf{reweighing} \citep{kamiran2012data}, and \textbf{disparate impact remover} \citep{feldman2015certifying} algorithms (for details on these algorithms, see Section \ref{lit-review-disc}). The results, summarized in \Cref{results-compared}, cover the three datasets: Adult, COMPAS, and German Credit.

Next, we evaluate the performance of FairUDT using the \(KL_{gain}\) and \(E_{gain}\) ratio splitting criteria. We also compare the results of FairUDT with the post-processing technique of discrimination-aware decision tree learning (DADT) proposed by \cite{kamiran2010discrimination}. Additionally, we present a performance comparison of various classifiers when using FairUDT on the Adult dataset.

As described in Section \ref{notation}, we consider two metrics for measuring algorithmic fairness in classification tasks: \textbf{demographic parity (DP)} and \textbf{average odds difference (AOD)}. We also report \textbf{balanced accuracy (BA)} and \textbf{accuracy (Acc)} for measuring the predictive performance of the classifiers after applying FairUDT.

\begin{table*}[t]
  \caption{Comparison of various pre-processing techniques with FairUDT (using $KL_{gain}$ as the splitting criterion and LR as the classifier). \textbf{Bold} indicates the best result, while \underline{underline} represents the second-best result}
  \label{results-compared}
  \centering
  \setlength{\tabcolsep}{4pt} 
  \linespread{1}\selectfont

  \begin{tabular}{llcc|cc}
  \toprule
    Dataset & Method                         & DP $ \downarrow$ & AOD $ \downarrow$ & BA $ \uparrow$ &  Acc $ \uparrow$\\ \midrule
    Adult   & Raw                            & -0.18                                  & -0.08                                       & 0.78 & 0.86
                                    \\
            &  Disparate Impact Remover       & \textbf{-0.02}                                  & 0.07                                        & \textbf{0.72}     & 0.74                             \\
            & Optimized Pre-processing       & 0.09                                   & 0.05                                        & 0.69    &  0.66                              \\
            & Reweighing                     & -0.09                                  & \underline{-0.03}                                       & \underline{0.71}    & 0.73                              \\
            & \textbf{Proposed Approach} \\
            & FairUDT + Raw Test Set & \underline{-0.07}                         & 0.04                              & 0.69    &  \underline{0.83}                     \\
            & FairUDT + Relabelled Test Set & \underline{-0.07}  & \textbf{0.00}                               & \underline{0.71}        & \textbf{0.86}                 \\
            \midrule
    COMPAS  & Raw                            & -0.17                                  & -0.15                                       & 0.67 & 0.67 
                                  \\
            & Disparate Impact Remover       & -0.14                                  & -0.10                                       & \textbf{0.69}     & \textbf{0.69}                             \\
            & Optimized Pre-processing       & -0.10                                  & -0.07                                       & \underline{0.68}     & \underline{0.68}                             \\
            & Reweighing                     & \underline{0.05}                                   & 0.10                                        & 0.63     & 0.63                             \\
            & \textbf{Proposed Approach}     \\ 
            & FairUDT + Raw Test Set & \textbf{0.00}                         & \underline{0.02}                               & 0.63    & 0.65                     \\
            & FairUDT + Relabelled Test Set  & \textbf{0.00}   & \textbf{0.00}                               & 0.63   & 0.66                      \\
            \midrule
    German Credit  & Raw                            & -0.10                                   & -0.09                                        & 0.68 & 0.76
                \\
            & Disparate Impact Remover       & \underline{-0.11}                                  & 0.81                                        & 0.54     & 0.56                             \\
            & Optimized Pre-processing       & -0.56                                  & -0.51                                       & 0.63      & 0.64                            \\
            & Reweighing                     & -0.22                                  & -0.19                                       & 0.65     &  0.62                             \\
            & \textbf{Proposed Approach} \\
            & FairUDT + Raw Test Set & \textbf{-0.03}                         & \underline{0.02}                               & \underline{0.67} &  \underline{0.76}                        \\
            & FairUDT + Relabelled Test Set & \textbf{-0.03}  & \textbf{-0.01}                               & \textbf{0.68}  & \textbf{0.77}                        \\ \bottomrule

  \end{tabular}
\end{table*}

\subsection{FairUDT vs. Key Related Pre-processing Techniques}

In \Cref{results-compared}, we report the BA, Acc, and fairness metrics for the related pre-processing techniques available in the existing literature and compare them with our best results, i.e., FairUDT with \(KL_{gain}\). The AOD, BA, and Acc metrics for our technique have been reported in two cases. In the first case, the test set was selected from the original data for evaluation, while in the second case, it was sampled from the relabeled dataset. The DP metric remains the same in both cases, as it is calculated solely based on the classifier's predictions. Additionally, raw results were reported for each dataset to compare the performance of the classifier (Logistic Regression, in this case) with and without applying data pre-processing techniques. The aim of this evaluation is to reduce discrimination compared to the raw dataset while preserving an acceptable level of accuracy.

For the Adult dataset, FairUDT achieves a perfect AOD of $0.00$ on its relabeled test set with a discrimination threshold of \(\sigma_t = 0.61\). However, the Disparate Impact Remover performs better in terms of DP, with FairUDT coming in second. Similarly, for BA, the Disparate Impact Remover ranks highest, followed by Reweighing and FairUDT. Notably, in terms of accuracy, FairUDT with the relabeled test set outperforms all other methods and maintains the same accuracy as the raw dataset.

FairUDT achieves state-of-the-art results for both DP and AOD on the COMPAS and German datasets. For COMPAS, with \(\sigma_t = 0.1\), FairUDT reaches a DP of $0.00$ on both the raw and relabeled test sets. In terms of AOD, FairUDT ranks highest and second-highest compared to previous pre-processing techniques. Although its performance in BA and accuracy is lower than that of the Disparate Impact Remover and Optimized Pre-processing, FairUDT overall achieves better results compared to other methods, balancing accuracy and discrimination effectively. Specifically, for the German dataset with \(\sigma_t = 1.64\), other approaches do not effectively mitigate discrimination to an acceptable level while achieving higher performance in BA and accuracy. FairUDT, on the other hand, maintains the same balanced accuracy as the raw dataset ($0.68$) while achieving the lowest discrimination values ($-0.03$ DP, $-0.01$ AOD) on both the raw and relabeled test sets. Additionally, FairUDT improves the prediction accuracy of the raw German Credit dataset, increasing it by $1\%$ on the relabeled test set (from $0.76$ to $0.77$).
\begin{figure}[ht]
\centering
\includegraphics[width=0.45\linewidth]{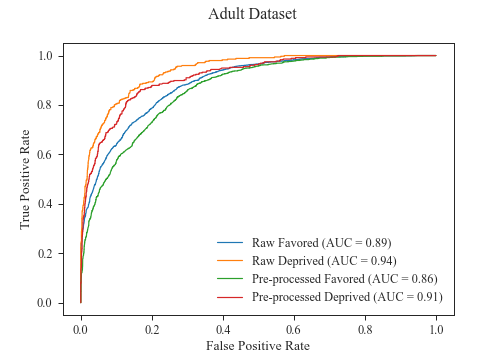}
\includegraphics[width=0.45\linewidth]{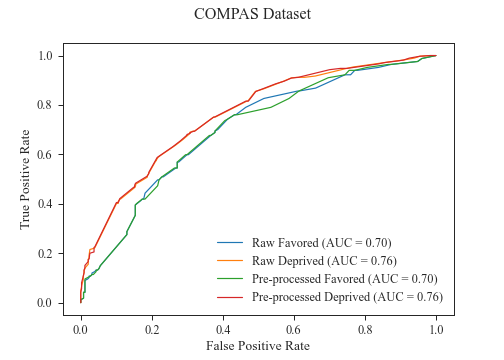}
\includegraphics[width=0.45\linewidth]{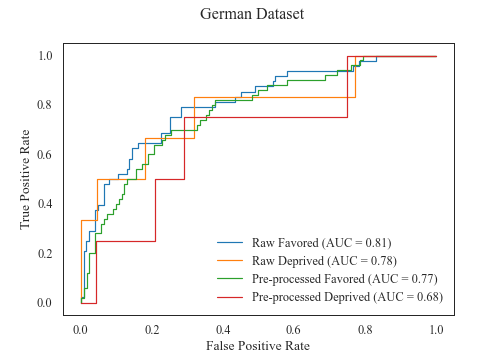}
\caption{ROC for all three datasets}
\label{roc-datasets}
\end{figure}

In \Cref{roc-datasets}, we have created ROC curves for all three datasets. The figure illustrates ROC curves for raw and pre-processed data, further divided into favored and deprived groups. Each curve is plotted for measuring the true positive rate (recall) and false positive rate at different classification thresholds using the LR classifier. As discussed in previous studies, data pre-processing for discrimination removal decreases the classifier's accuracy and hence, its AUC. The resultant AUC curves shown in \Cref{roc-datasets} after pre-processing have lower AUC values compared to the raw datasets. However, the ratio between favored and deprived groups remains the same before and after pre-processing, resulting in the same classification thresholds as the raw datasets. This shows that our pre-processing technique preserves the true positive rate and false positive rate of the original dataset and is hence independent of the classifier used for classification.

\begin{figure}[ht]
\centering
\includegraphics[width=1\linewidth]{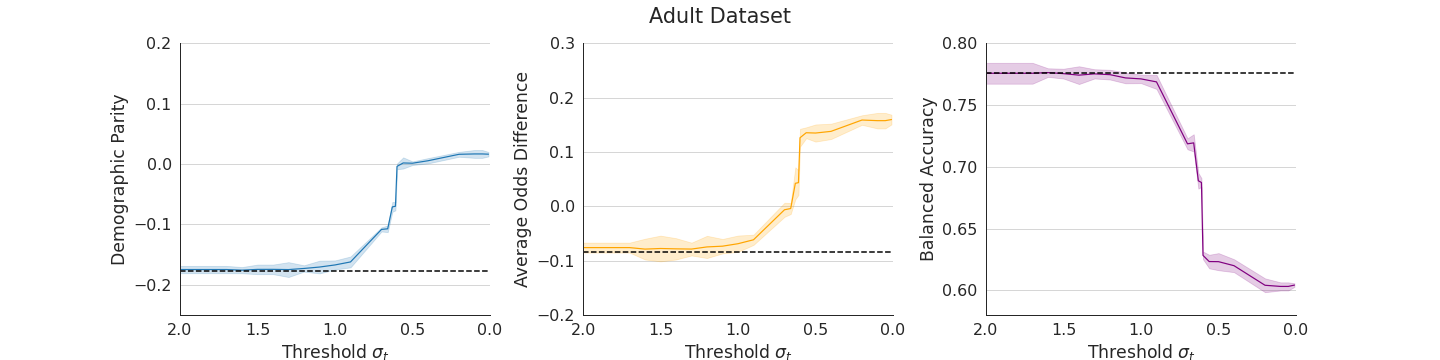}
\includegraphics[width=1\linewidth]{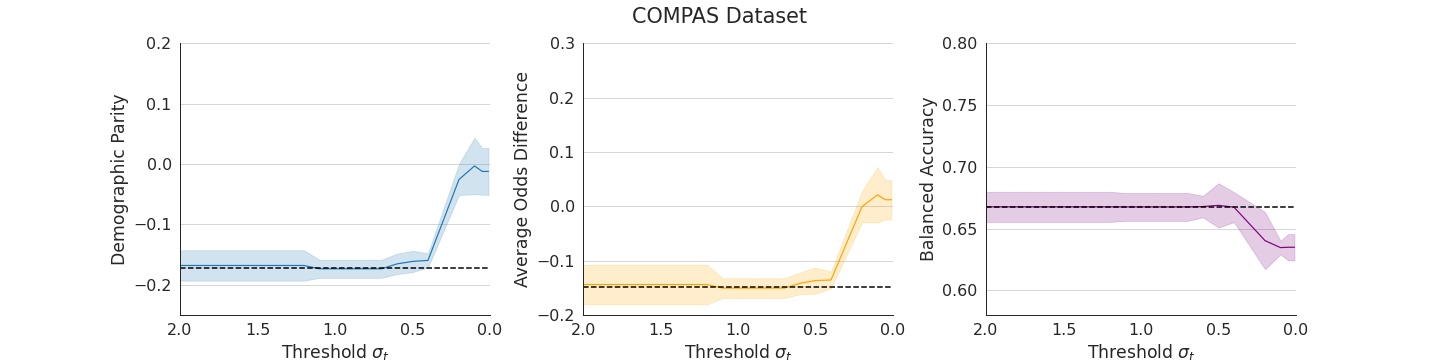}
\includegraphics[width=1\linewidth]{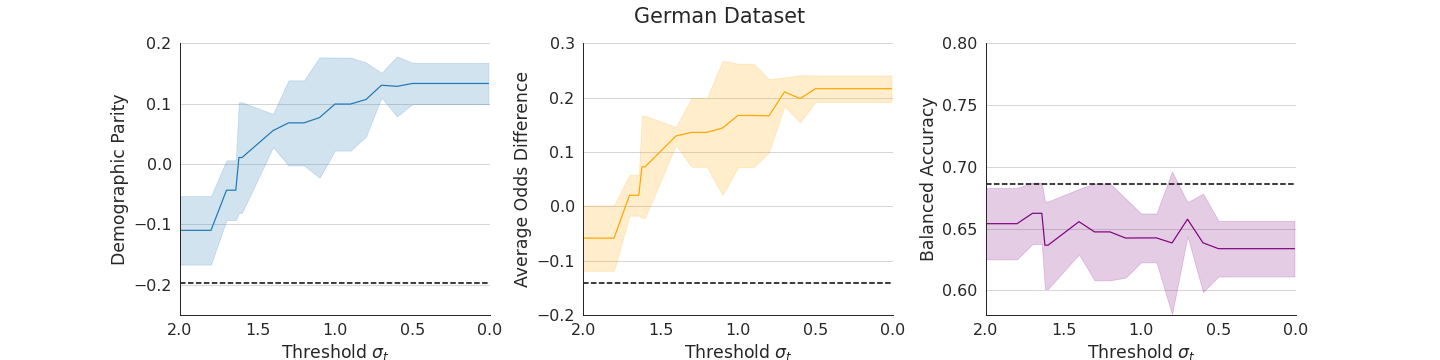}
\caption{LR classification results for the three datasets after pre-processing data at different thresholds of $\sigma_t$. Error bands show the standard deviation of different results from 10-fold cross-validation. The black dashed line shows the corresponding metric after evaluating LR on the original dataset. The test set is sampled from raw data.}
\label{kl-tree-results}
\end{figure}

In \Cref{results-compared}, all the datasets are pre-processed using the optimal discrimination thresholds of \(\sigma_t\). This is illustrated in \Cref{kl-tree-results}, where we show how the DP, AOD, and BA metrics change with different values of the discrimination threshold \(\sigma_t\) across all three datasets using the LR classifier. Results of these metrics on the raw test sets are shown with a black dashed line on each corresponding plot. From the plots, it can be seen that the fairness metrics of DP and AOD can be successfully tuned using FairUDT. As shown, the optimal value of $0.00$ discrimination for these metrics can be achieved at certain thresholds of \(\sigma_t\). For the Adult and COMPAS datasets, these metrics change gradually when tuning \(\sigma_t\) from $2.0$ to $0$, while an abrupt change is seen around the values of $0.61$ (Adult) and $0.1$ (COMPAS). This is because optimal subgroups at leaves are formed around these discrimination values \(disc_l\). In contrast, this behavior is not observed for the German Credit dataset, where gradual changes in the fairness metrics are seen with varying \(\sigma_t\). Additionally, the German Credit dataset achieves nearly $0.00$ DP and AOD around the discrimination threshold of \(\sigma_t = 1.64\). A possible explanation for this is that the German dataset contains more features (20), each with numerous unique values, leading to the formation of more fine-grained and approximately equal-sized leaves. We conclude that our approach is most effective when the original dataset has a large number of features.

As expected, balanced accuracy (BA) decreases as more leaves are relabeled by moving \(\sigma_t\) towards 0. For the Adult dataset with the raw test set, this deterioration in BA is notably more significant compared to the other datasets. In the experiments, it was observed that at lower values of the discrimination threshold \(\sigma_t\), relabeling for the Adult dataset primarily involved leaf demotions. This resulted in a larger proportion of negative-class (\(\leq 50k\)) samples in the pre-processed dataset. Consequently, the classifier (LR) trained on this dataset became biased towards the negative class, leading to reduced balanced accuracy. Therefore, we emphasize the importance of tuning \(\sigma_t\) to an appropriate value according to the characteristics of the underlying dataset, in order to find an acceptable discrimination-accuracy tradeoff.

In general, this phenomenon applies to any probabilistic classifier (LR, SVM, or DT) when using the leaf relabeling technique, as the promotion and demotion of instances at the leaves affect the redistribution of class labels. Hence, relabeling may enhance fairness metrics but negatively impact balanced accuracy, particularly if the process disproportionately affects instances of one class. A performance comparison of various classifiers on the pre-processed Adult dataset (\(\sigma_t = 0.61\)) is presented in \Cref{classifier_comparison}.

\begin{table}[ht]

\caption{Performance comparison of various classifiers, including Logistic Regression (LR), Support Vector Machine (SVM), and Decision Tree (DT), after applying FairUDT on the Adult dataset. Results are reported on Raw test set.}
\label{classifier_comparison}
\centering
\linespread{1}\selectfont
\begin{tabular}{lll|ll}
\toprule
\multicolumn{1}{c}{Classifier} & DP $\downarrow$   & AOD $\downarrow$  & BA $\uparrow$  & Acc $\uparrow$ \\ \midrule
LR                             & -0.07 & 0.04 & 0.69 & 0.83 \\ 
SVM                            & -0.06 & 0.07  & 0.67 & 0.82 \\ 
DT                             & -0.08 & 0.01  & 0.66 & 0.82 \\ \bottomrule
\end{tabular}
\end{table}


\subsection{$KL_{gain}$ vs. ${E}_{gain}$}

 In this work, all experiments utilized FairUDT with $KL_{gain}$, as it consistently outperformed ${E}_{gain}$ in terms of fairness metrics across all datasets and classifiers used. 
 A performance comparison between $KL_{gain}$ at $\sigma_t=0.61$ and ${E}_{gain}$ at $\sigma_t=0.01$ on the Adult dataset using the LR classifier is presented in \Cref{klgain_egain}. The optimal values of $\sigma_t$ were selected to report the results for both information gain criteria. From the results, we see that ${E}_{gain}$ performs better in terms of Acc and BA, but gives poor performance on fairness metrics. The reason behind better performance of the $KL_{gain}$ based splitting criterion in fairness metrics can be attributed to its asymmetric nature. In our model, KL-divergence identifies splits where the distribution of the favored dataset deviates from that of the deprived dataset. Specifically, it measures the information gain when approximating the favored distribution with the deprived distribution, a gain that cannot be achieved when approximating the deprived distribution from the favored one. This directionality is crucial, highlighting why symmetric measures like Euclidean distance are inadequate in our context, as they fail to convey this directional information.

 \begin{table}[ht]

\caption{Performance comparison of FairUDT with ${KL}_{gain}$ and ${E}_{gain}$ ratio splitting criteria (taking Adult dataset and LR classifier). Results are reported on Raw test set.}
\label{klgain_egain}
\centering
\linespread{1}\selectfont
\begin{tabular}{lcc|cc}
\toprule
\multicolumn{1}{c}{Classifier} & DP $\downarrow$   & AOD $\downarrow$   & BA $\uparrow$   & Acc $\uparrow$  \\ \toprule
FairUDT + ${KL}_{gain}$             & -0.07 & 0.04  & 0.69 & 0.83 \\ 
FairUDT + ${E}_{gain}$              & -0.16 & -0.08 & 0.76 & 0.85 \\ \bottomrule
\end{tabular}
\end{table}

\subsection{FairUDT vs. DADT}

Given that FairUDT is motivated by the Discrimination-aware Decision Tree learning (DADT) algorithm proposed by \cite{kamiran2010discrimination}, we have presented a comparative analysis of FairUDT with the post-processing DADT technique in \Cref{kamiran-results}. Specifically, we use the $IGC+IGS\_Relab$ method from DADT to present the comparison results with FairUDT + $KL_{gain}$ ($\sigma_t = 0.61$). In the table, Raw denotes the results obtained from the original Adult dataset using a Decision Tree (DT) classifier. To maintain consistency, we have also employed the DT classifier to evaluate the predictive performance of the data pre-processed using FairUDT. The results indicate that while DADT slightly outperforms FairUDT in terms of demographic parity (DP), the average odds difference (AOD) achieved by our approach is significantly closer to $0.00$. Regarding predictive performance, FairUDT demonstrates competitive balanced accuracy (BA) and accuracy compared to DADT.

\begin{table} [ht]

    \caption{Comparison of FairUDT (using  $KL_{gain}$ as the splitting criterion and DT as the classifier) with DADT technique on Adult dataset. FairUDT results are reported on Raw test set. We report fairness metrics as unsigned.}
    \label{kamiran-results}
    \centering
    \linespread{1}\selectfont
\begin{tabular}{lcc|cc}
\toprule
Method  & DP $\downarrow$   & AOD $\downarrow$  & BA $\uparrow$ & Acc $\uparrow$   \\ \midrule
Raw     & 0.17 & 0.09 & 0.74     &  0.84 \\ 

DADT    & \textbf{0.08} & 0.04 & 0.66     & 0.82 \\
FairUDT & 0.09 & \textbf{0.01} & \textbf{0.69}     & \textbf{0.83}
   \\ \bottomrule
\end{tabular}
\end{table}


\subsection{FairUDT as an Interpretable Decision Tree}

\Cref{tab:interpretabilityFairUDT} presents a comparative analysis of interpretability of FairUDT and the conventional decision tree. The assessment of interpretability is based on four metrics: the total number of nodes (encompassing both branches and leaves), sparsity, total tree depth, and simulatability \citep{aghaei2023optimalfair, rudin2022interpretable, molnar2022}. These metrics are evaluated across all the three datasets: Adult, COMPAS, and German Credit. From the table, we can see that FairUDT trees are shallower and sparser, resulting in higher simulatability. This is particularly evident in the COMPAS dataset, where simulatability is rated as `High' by human evaluators. In contrast, decision trees tend to exhibit a greater number of nodes and increased depth, leading to lower simulatability, as seen in the Adult dataset, where simulatability is rated as `Low'. This comparison highlights that FairUDT maintains or enhances interpretability across various datasets compared to traditional decision trees. The details pertaining to training and parameters of decision trees, and measurement of simulatability can be seen in \ref{InterpretableDTAppendix}.

\begin{table}

    \centering
    \smaller
    \setlength{\tabcolsep}{3pt} 
\caption{Comparison of the interpretability of FairUDT and Decision Trees using four measures: number of nodes, sparsity, depth, and simulatability.}
\label{tab:interpretabilityFairUDT}
    \begin{tabular}{c|l|ccc}
    \toprule
       Method & Measure &  Adult&  COMPAS&  German Credit\\
         \midrule
        FairUDT & Number of nodes& 3813 & 114 & 492 \\
        & Sparsity& 1619 & 42 & 192 \\
        & Depth& 10 & 7 & 11 \\
        & Simulatability& Medium & High & Medium\\
        \midrule
        Decision Tree &  Number of nodes& 6077 & 297 & 255\\
         & Sparsity& 3039 & 149 & 128 \\
         & Depth& 38 & 10 & 15 \\
        & Simulatability& Low & Medium-High & Medium \\
 \bottomrule
    \end{tabular}

\end{table}

\section{Discussion}
In this section, we discuss the operational and societal implications of FairUDT. We also highlight the limitations of our proposed method and suggest possible future directions to address these challenges.

\subsection{Practical Impact}
FairUDT has numerous practical applications across industries by ensuring subgroup-level fairness in decision-making. For example, in HR and recruitment, it can help organizations comply with anti-discrimination laws by eliminating bias in candidate selection. Similarly, in financial services, it ensures equitable credit scoring and loan approvals while maintaining compliance with regulations.

An important feature of FairUDT is to control discrimination at subgroup level by fine tuning a hyperparameter in the pre-processing phase. This would be beneficial in order to meet different business needs and regulation constraints. The use of FairUDT will also help organizations comply with anti-discrimination laws and regulations, fostering trust from customers, employees, and the public. By promoting ethical AI practices and maintaining the interpretability of decision trees, FairUDT ensures understandable and trustworthy decisions. Additionally, as depicted in Figure \ref{kl-tree-results}, the proposed model balances fairness with predictive accuracy, leading to effective and equitable outcomes, thereby enhancing the overall utility of decision-making processes.


\subsection{Theoretical Impact }
The theoretical impact of FairUDT lies in its novel approach to fairness-aware decision tree learning. To the best of our knowledge, it is the first framework to integrate discrimination-aware strategies with uplift modeling. Additionally, the proposed leaf relabeling strategy enables finer control over subgroup-level discrimination detection, advancing fairness methodologies. By functioning as a generalized data pre-processing tool independent of the classifier, FairUDT opens opportunities for seamless integration with other fairness techniques, such as in-processing algorithms and post-processing adjustments, paving the way for holistic fairness solutions.

FairUDT can advance current fairness research and lay a foundation for future exploration in uplift modeling for discrimination detection and fairness-aware modeling.

\subsection{Societal Impact}
FairUDT can contribute positively to society by addressing biases in automated decision-making, promoting fairness, and fostering inclusivity in areas like employment, finance, and healthcare. By ensuring decisions are equitable, it can reduce systemic discrimination and build trust in AI systems. However, there are potential risks: improper tuning of the model’s parameters could inadvertently amplify biases, and reliance on FairUDT without comprehensive checks may lead to overconfidence in fairness outcomes. These challenges highlight the need for careful implementation and regular audits to ensure its social benefits outweigh the risks.


\subsection{Limitations}

FairUDT operates by generating a tree to its maximum depth to capture highly biased instances at the leaves. While this approach can be effective for smaller datasets (such as those with fewer than 100k records), it may lead to overfitting and computationally complex trees in large-scale or streaming data settings, thereby impacting FairUDT's performance. However, we emphasize two key points: 1) training classifiers for socially sensitive tasks is not time-constrained, as the main focus of such applications is on ensuring fairness \cite{aghaei2019learning}, and 2) under certain sparsity constraints, decision trees can perform well on datasets with large number of samples and features \cite{Klusowski02012023}. Additionally, FairUDT's current formulation requires binary class labels and binary sensitive attributes, which could pose a significant challenge in real-world scenarios where multiple classes and multiple sensitive attributes exist. Moreover, FairUDT cannot handle concept drift in online data streams. To address this, we recommend conducting frequent retrainings and comparing model performance to select the best one based on accuracy and fairness metrics.

Studies on model transparency \citep{aghaei2023optimalfair, carvalho2019machine} have suggested that more interpretable models, such as decision trees, perform slightly worse than complex models such as random forests (RF) and neural networks (NN). Therefore, in domains where achieving higher accuracy is a priority, compromising on transparency might be a necessary trade-off.

\subsection{Future Directions}

To date, discrimination is a subjective term, and there is no consensus on what constitutes discrimination. In this study, we aligned our efforts with the U.S. Equal Employment Opportunity Commission (EEOC) \cite{usgovteeoc}. However, the broad applicability of FairUDT allows for further extensions. In the future, we aim to enhance FairUDT to address intersectional or compound discrimination through multi-treatment uplift modeling. Another important aspect to improve our model's performance would be to incorporate FairUDT into ensemble models like Random Forests and XGBoost. Additionally, since our pre-processing approach involves a non-convex optimization problem, future work will explore convex optimization pre-processing frameworks based on uplift modeling.

\section{Conclusion}

In this paper, we introduce FairUDT, a data pre-processing approach designed to generate discrimination-free datasets. To the best of our knowledge, this is the first work to apply the concept of uplift modeling to discrimination identification. Specifically, we present a decision tree-based discrimination identifier that leverages uplift modeling to discover discriminatory subgroups within the data. Additionally, we introduce a modified leaf relabeling criterion to remove discrimination at the tree leaves. Our results demonstrated that FairUDT is applicable to any real-world dataset without requiring feature engineering and produces competitive results compared to key related pre-processing approaches. In terms of popular fairness metrics and predictive performance, the results of FairUDT confirm the effectiveness of our approach. We also demonstrate that FairUDT is inherently interpretable and can be used in discrimination detection tasks. Our proposed pre-processing method preserves utility in decision-making and ensures group fairness at a granular level. Moreover, through probabilistic descriptions, our method connects to the broader literature on statistical learning and information theory.

\section*{Funding}

No funding was received for conducting this study.

\section*{CRediT authorship contribution statement}
Conceptualization; Anam Zahid, Abdur Rehman Ali, Rai Shahnawaz, Shaina Raza; Data curation: Anam Zahid, Abdur Rehman Ali; Formal analysis: Anam Zahid, Abdur Rehman Ali, Rai Shahnawaz; Methodology: Anam Zahid, Abdur Rehman Ali, Rai Shahnawaz; Project administration: Faisal Kamiran, Asim Karim, Shaina Raza; Supervision: Faisal Kamiran, Asim Karim, Shaina Raza; Validation: Faisal Kamiran, Asim Karim, Shaina Raza; Visualization: Abdur Rehman Ali; Writing – original draft: Anam Zahid, Rai Shahnawaz; Writing – review \& editing: Faisal Kamiran, Asim Karim, Shaina Raza

\section*{Declaration of competing interest}
The authors declare that they have no known competing financial
interests or personal relationships that could have appeared to influence
the work reported in this paper.






\appendix
\section{Notations and Definitions}
\label{notationsAppendix}
We are given a dataset \((X, S, Y)\) consisting of \(N\) number of records. Here, \(X\) represents the elements of non-sensitive attributes. Without loss of generality, assume that \(S = \{S_F, S_D\}\) is a binary sensitive attribute such as gender or race, and the outcome \(Y\) is a binary variable. The dataset is further split into two groups; favored \((X, S_F, Y)\) and deprived \((X, S_D, Y)\), based on the values of \(S\). Further, \(N^F\) denotes the number of records in the favored group, and \(N^D\) indicates the number of records in the deprived group respectively (i.e., \(N = N^F + N^D\)). The class variable \(Y=\{Y^+, Y^-\}\) has \(y\) as a particular label among a finite number of possible outcomes. Hence, all involved probability distributions are discrete. The probabilities for the favored and deprived groups are represented by \(P^F\) and \(P^D\), respectively, while \(P^F(Y)\) and \(P^D(Y)\) denote the probabilities of the class attribute within the favored and deprived groups.

In standard decision trees, there are separate testing techniques for categorical and numerical attributes. In our context, there is a single split test \(A\) for each categorical attribute which produces a finite number of outcomes \(a \in A\) based on distinct values. For numerical attributes, data discretization has been performed; both favored and deprived groups are combined into a single dataset for this purpose. For a specific test \(A\), conditional probabilities over favored and deprived groups would have the form \(P^F(Y|A)\) and \(P^D(Y|A)\) separately. Here, \(P^F(Y|a)\) and \(P^D(Y|a)\) define the estimated probability distributions for the favored and deprived groups respectively, for the specific outcome \(A=a\).

Next, we define the two fairness criteria implemented in FairUDT: demographic parity and average odds difference. Additionally, we introduce the mathematical notation for balanced accuracy, the metric used alongside accuracy to evaluate the performance of the data classifier.

\textbf{Demographic parity (DP):} Demographic parity (DP), also known as total variation (TV), statistical parity, or risk difference, is defined as measuring the discrepancy between predicting a positive outcome \(\hat{Y} = Y^+\) under \(S = S_D\) and \(S = S_F\). Formally, it can be written as:
\begin{equation*}
    P(\hat{Y} = Y^+ | S = S_F) = P(\hat{Y} = Y^+ | S = S_D)
\end{equation*}

In real life, achieving this equality constraint is very difficult. Hence, a more relaxed criterion known as Difference in Demographic Parity (DDP) is defined:
\begin{equation*}
    DDP = P(\hat{Y} = Y^+ | S = S_F) - P(\hat{Y} = Y^+ | S = S_D)
\end{equation*}

A classifier is said to be fair if the value of its DDP is 0, i.e., parity across all groups. The highest and lowest values of 1 and -1, respectively, signify complete unfairness. For clarity and consistency, we have used the term demographic parity (DP) throughout our paper to refer to DDP.

\textbf{Average odds difference (AOD):} Before defining average odds difference (AOD), it is imperative to define equality of odds (EOD), since AOD is a relaxed version of it. EOD is satisfied when both the True Positive Rate (TPR) and False Positive Rate (FPR) are independent of the value of the sensitive attribute. It is defined as follows:
\begin{equation*}
\begin{split}
   & TPR: P(\hat{Y} = Y^+ | S = S_F, Y = Y^+ ) - P(\hat{Y} = Y^+ | S = S_D, Y = Y^+) = \\
  & FPR: P(\hat{Y} = Y^+ | S = S_F, Y = Y^- ) - P(\hat{Y} = Y^+ | S = S_D, Y = Y^-)
\end{split}
\end{equation*}
In its general form, EOD can be written as:
\begin{equation*}
\begin{split}
   & P(\hat{Y} = Y^+ | S = S_F, Y = y) = P(\hat{Y} = Y^+ | S = S_D, Y = y) \\
   & = P(\hat{Y} = Y^+ | Y = y) \hspace{0.5cm} \forall y \in \{+, -\}
\end{split}
\end{equation*}
Similarly, AOD returns the mean value of the difference between TPR and FPR for the favored and deprived groups. It can be specified as:
\begin{equation*}
    \frac{[TPR_{S = S_D} - TPR_{S = S_F}] + [FPR_{S = S_D} - FPR_{S = S_F}]}{2}
\end{equation*}
An AOD of 0 indicates a perfect EOD.

\textbf{Balanced accuracy (BA):} To measure the performance of the classifier, we calculate balanced accuracy (BA) instead of measuring accuracy alone. BA measures the average accuracy across all classes. Mathematically, it is calculated as the arithmetic mean of TPR and TNR (True Negative Rate):
\begin{equation*}
    BA = \frac{TPR + TNR}{2}
\end{equation*}

Balanced accuracy provides a fairer assessment of model performance compared to traditional accuracy, particularly when classes are imbalanced, as it considers both the \(Y^+\) and \(Y^-\) class performance.

\section{KL-Divergence Propositions}
\label{KLAxiomAppendix}
\paragraph{Proposition 3.1:} $KL_{gain}$ will be minimum iff favored and deprived class distributions are identical for all outcomes of a test A (i.e. $P^{F}(Y|a)=P^{D}(Y|a)$ ). \\
It is driven by the fact that gain value should be maximized in proportion to the divergence achieved between favored and deprived class distributions. It is an accepted property of Kullback-leibeler divergence that it would be always greater than or equal to zero $KL(P^{F}|P^{D})\geq0$, known as Gibbs inequality \citep{falk1970inequalities}. Hence, for identical $P^{F}$ and $P^{D}$ distributions it would have zero value and consolidating it with the fact that unconditional (second) term of $KL_{gain}$ is independent of the test proves the claim.

\paragraph{Proposition 3.2:}$KL_{gain}$  would be zero if A is statistically independent of Y for both favored and deprived distributions i.e. $P^{F}(Y|a) = P^{F}(Y)$ and $P^D(Y |a) =
P^D(Y)$. \\
It will ensure that tests with no information gain for class are not selected for the split as in standard decision trees.  Using assumption from the statement that A is statistically independent of Y, we can replace $f_y(a)$ with $f_y$ (class probabilities test independence for favored distribution) and $d_y(a)$ with $d_y$ (class probabilities test independence for favored distribution) in \Cref{KLdivergenceGain}, it will equalize both the terms involved in the equation eventually making $KL_{gain}(A)$ equals zero. However, class
distributions after the split can be more similar than before which will lead to negative values for splitting criterion. It implies that there can be split tests even worse than the independent split. 

\paragraph{Proposition 3.3:}$KL_{gain}$ reduces to entropy gain when deprived dataset is absent.

\paragraph{Theorem 3.1}
$KL_{gain}$ test does satisfy the stated propositions 3.1-3.3. Proof for proposition 3.3 can be found in \cite{rzepakowski2010decision}, and reduction of $KL_{gain}$ to entropy gain can be seen in \cite{quinlan1986induction}.

\section{Squared Euclidean Distance Propositions}
\label{EucledianAxiomAppendix}

\paragraph{Proposition 3.5:} $E_{gain}$ will be minimum iff favored and deprived class distributions are same for all outcomes of a test A. \\
It is a well known and understandable property of Euclidean distance to be always greater than or equal to zero as it is a square of difference between probability distributions values. Further, an argument similar to the one used in proof of proposition 3.1 proves the claim.

\paragraph{Proposition 3.6:}$E_{gain}$  would be zero if A is statistically independent of Y for both favored and deprived datasets.\\
Deriving from the assumption, we can replace $f_y(a)$ with $f_y$ and $d_y(a)$ with $d_y$ in \Cref{EuclideanGain}, as a consequence it will balance the two terms involved, hence making $E_{gain}(A)$ equals zero.

\paragraph{Proposition 3.7:\label{Axiom3.7}}$E_{gain}$ reduces to $GINI_{gain}$ if either favored or deprived group is missing. \\
$E_{gain}$ reduction to $GINI{gain}$ \citep{brieman1984classification} is demonstrated in \cite{rzepakowski2010decision}, when either of the group is absent.

\paragraph{Theorem 3.2}
$E_{gain}(A)$ verifies the propositions 3.5-3.7 as proved above.

\section{Interpretable Decision Trees}

\label{InterpretableDTAppendix}

In this section, we elucidate our evaluation of the simulatability of FairUDT. To measure simulatability, we first identify subgroups at the leaves that exhibit high discriminatory behavior. A subgroup is defined by a specific set of attributes and their values, representing the input samples for that subgroup. This identification is accomplished by selecting a single tree leaf and tracing the path from the root node to that leaf. Several discriminatory subgroups for the Adult dataset are shown in \ref{adult-explainability}.

For instance, the first subgroup in the table contains seven instances, six of which belong to favored individuals (male) from the positive class ($m^+$), while only one instance belongs to the deprived group (female) from the negative class ($f^-$). This subgroup exhibits the highest discrimination value ($2.0$) in our dataset. Similarly, the fifth subgroup shows a discrimination value of $1.1$, which can be explained by looking at the gender distribution across the positive and negative classes within the subgroup.

We collected 30 subgroups containing discrimination, as well as subgroups that are free from discrimination, from the Adult, COMPAS, and German Credit datasets. Subsequently, we asked four computer science graduates to assign each subgroup a discrimination value (from 0 to 2), and categorize it on a scale from `High' to `Low' based on their understanding of the attributes and their values within that subgroup. The simulatability results reported in \Cref{tab:interpretabilityFairUDT} represent the average results for each dataset.

We then trained the decision tree using the default parameters from the scikit-learn library \cite{scikit-learn}, with the exception of setting the minimum number of samples in the leaf node to 3 to avoid overfitting. As discussed above, the same procedure was repeated to determine the simulatability of the decision tree for each dataset. 
\begin{table}[htbp]
  \caption{Some discriminatory subgroups along with their discrimination value $(disc_l)$,  detected by KL-gain tree on Adult dataset. Here, $S = gender$, $S_F = male$ and, $S_D = female $ }
  \label{adult-explainability}
  \small
    \begin{tabular*}{\columnwidth}{clccc@{}}
    \cmidrule(l){1-5}
      No. & Subgroup & $m^+: m^-$  & $f^+:f^-$                                                                 &  $disc_l$ \\
      \cmidrule(l){1-5} 
    1 & \begin{tabular}[c]{@{}l@{}}occupation=Craft-repair\\ capital\_gain\textless{}5119.0\\ race=White\\ education=Masters\\ hours\_per\_week\textgreater{}40.5\\ native\_country=United-States\\ workclass=Private\end{tabular}               & 6 : 0 & 0 : 1 & 2.0                  \\
      &                                                         &                   \\ 
    2 & \begin{tabular}[c]{@{}l@{}}occupation=Exec-managerial\\ workclass=Self-emp-inc\\ capital\_loss=1537-1881\\ relationship=Not-in-family\\ age=40-50\end{tabular}                                                                                                               & 1 : 0 & 0 : 1 & 2.0                  \\
      &                                                                 &                      \\
    3 & \begin{tabular}[c]{@{}l@{}}occupation=Craft-repair\\ capital\_gain\textless{}5119.0\\ race=Other\\ educational\_num=12.0-12.1\\ hours\_per\_week=37.5-40.5\end{tabular}                                                                                                & 1 : 0 & 0 : 1      & 2.0                  \\
      &                                                                                                                                                                                                                                                                                &                      \\
    4 & \begin{tabular}[c]{@{}l@{}}occupation=Exec-managerial\\ workclass=Self-emp-not-inc\\ educational\_num\textgreater{}12.1\\ marital\_status=Never-married\\ capital\_gain\textless{}5119.0\\ age=40.0-50.0\\ hours\_per\_week\textgreater{}40.5\\ education=Masters\end{tabular} & 1 : 0 & 0 : 1 & 2.0                  \\
      &                                                                                                                                                                                                                                                                                &                      \\
    5 & \begin{tabular}[c]{@{}l@{}}occupation=Craft-repair\\ capital\_gain\textless{}5119.0\\ race=Black\\ workclass=Private\\ hours\_per\_week\textgreater{}40.5\\ marital\_status=Married-civ-spouse\end{tabular}                                                                & 11 : 9 & 0 : 2   & 1.1                  \\ \cmidrule(l){1-5} 
    \end{tabular*}%
\end{table}

\bibliographystyle{elsarticle-num-names} 
\bibliography{mybibfile}



\end{document}